\documentclass[journal]{IEEEtran}

\usepackage{graphicx}
\usepackage{amsmath}
\usepackage{bm}
\usepackage{cite}

\begin{document}

\title{Temporal Attribute-Appearance Learning Network for Video-based Person Re-Identification}

\author{Jiawei~Liu,~Xierong~Zhu and Zheng-Jun~Zha~\IEEEmembership{Member,~IEEE}
	\thanks{Jiawei Liu, Xierong Zhu and Zheng-Jun Zha are with the School of
		Information Science and Technology, University of Science and Technology
		of China, Hefei, 230027, China. Email: jwliu6@ustc.edu.cn}
}

%
%

\markboth{IEEE Trans. ~X, ~Vol.~X, No.~X, August~2020}%
{Jiawei Liu \MakeLowercase{\textit{et al.}}: Temporal Attribute-Appearance Learning Network for Video-based Person Re-Identification}
%

\maketitle

\begin{abstract}
Video-based person re-identification aims to match a specific pedestrian in surveillance videos across different time and locations. Human attributes and appearance are complementary to each other, both of them contribute to pedestrian matching. In this work, we propose a novel Temporal Attribute-Appearance Learning Network (TALNet) for video-based person re-identification. TALNet simultaneously exploits human attributes and appearance to learn comprehensive and effective pedestrian representations from videos. It explores hard visual attention and temporal-semantic context for attributes, and spatial-temporal dependencies among body parts for appearance, to boost the learning of them. Specifically, an attribute branch network is proposed with a spatial attention block and a temporal-semantic context block for learning robust attribute representation. The spatial attention block focuses the network on corresponding regions within video frames related to each attribute, the temporal-semantic context block learns both the temporal context for each attribute across video frames and the semantic context among attributes in each video frame. The appearance branch network is designed to learn effective appearance representation from both whole body and body parts with spatial-temporal dependencies among them. TALNet leverages the complementation between attribute and appearance representations, and jointly optimizes them by multi-task learning fashion. Moreover, we annotate ID-level attributes for each pedestrian in the two commonly used video datasets. Extensive experiments on these datasets, have verified the superiority of TALNet over state-of-the-art methods.
\end{abstract}


\IEEEpeerreviewmaketitle

\section{Introduction}
%
%
%
%
\IEEEPARstart{P}{erson re-identification} is an important technology to find a target pedestrian across different, non-overlapping views \cite{wang2018mancs,chen2018improving,karianakis2018reinforced,liu2018ca3net}. It has drawn increasing attention in the computer vision community during recent years, as it plays a significant role in various critical surveillance applications, such as pedestrian tracking, behavior analysis and crowd counting \cite{liu2016multi} \textit{etc}. The surge of deep learning technique has been reflected in the task of person re-identification as well, with the exciting progress on many benchmark datasets. However, it remains to be challenging in real-world unconstrained scenes \cite{chen2018temporal}, result from cluttered background, partial occlusion, heavy illumination changes, error detection, viewpoint variations as well as non-rigid deformation of human body, \textit{etc}.

\begin{figure}[!t]
	\centering
	\includegraphics[width=0.5\textwidth]{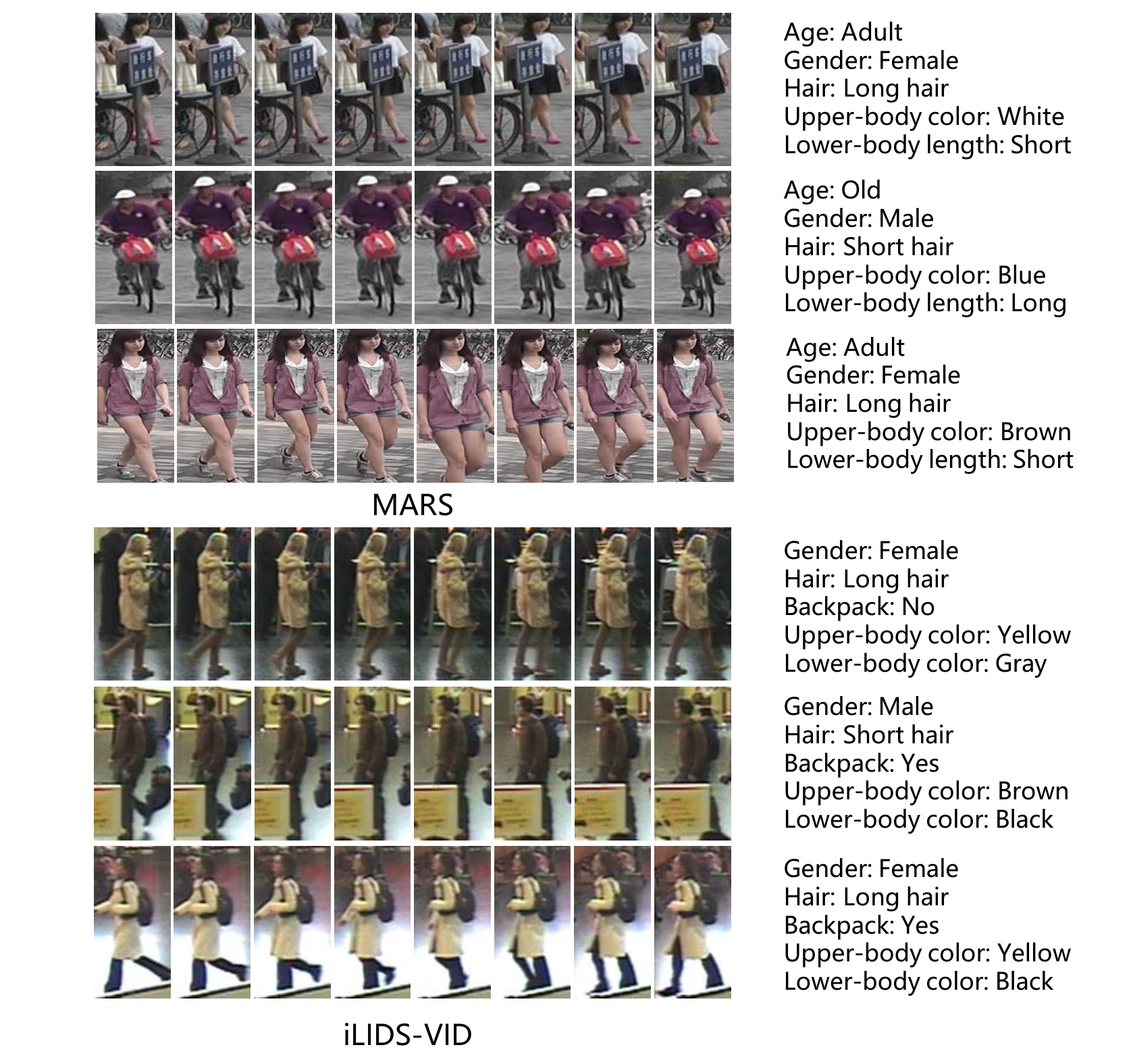}
	\caption{Example video sequences with several corresponding attributes in MARS and iLIDS-VID datasets. We sample eight frames for each video sequence.} 
	\label{example}
\end{figure}

Most existing approaches recognize the identities of pedestrians in the static image setting, which mainly focus on learning discriminative appearance feature of pedestrian \cite{chen2016similarity, liao2015person, varior2016learning} or deriving appropriate distance metric \cite{zheng2011person, cheng2016person, koestinger2012large} for feature matching. In parallel with the impressive progress of image-based person re-identification, video-based person re-identification has recently attracted a significant amount of attention. It aims to search for a video sequence of a particular pedestrian from a gallery of video set. Fig.\ref{example} shows some video examples of six pedestrians in the two benchmark datasets, MARS \cite{zheng2016mars} and iLIDS-VID \cite{wang2014person}. Different from an image containing limited information of pedestrian appearance, a video sequence contains richer motion information \cite{jiao20183d, liu2018dense}, which is complementary to pedestrian appearance. Moreover, a video sequence captures abundant temporal contextual information in a long span of time, presenting human appearance under different pose and viewpoint variations, which could provide crucial knowledge to alleviate visual ambiguity.

The key for video-based person re-identification is to abstract effective video-level representation. Some preliminary existing methods are just the straightforward extension solution of person re-identification in images. These methods extract frame-level appearance representation for each video frame independently, followed by temporal feature aggregation function (\textit{e.g}, different pooling operations and recurrent neural networks) to generate a sequence-level appearance representation \cite{mclaughlin2016recurrent, yan2016person, liu2018video}. They view different video frames of a sequence with equally important when aggregating frame-level features, leading to unsatisfactory performance. As shown in Fig.\ref{example}, a video sequence presents occlusion, viewpoint and pose variations \textit{etc} within different video frames. Consequently, different video frames provide visual cues with varying degrees of importance for matching pedestrians. Based on the above observation, a few of methods \cite{zhou2017see, chen2018temporal,ouyang2019video} build temporal attention modules to explore the importance scores for each frame, and then concentrate the models on relevant frames. Nevertheless, these methods only abstract global appearance representation from the whole body, while neglect fine-grained local visual cues from body parts.


Human attributes, \textit{e.g.}, \textit{short hair}, \textit{long dress} and \textit{red shoes}, are one kind of middle-level semantic descriptors for pedestrians, which provides vital information to match pedestrians. In contrast with visual appearance, human attributes have great ability of robustness against the aforementioned challenges \cite{lin2017improving}. With regard to pedestrians sharing similar appearance or a pedestrian having large appearance discrepancy, appearance representation usually fails to recognize pedestrians. Nevertheless, human attributes, as middle-level semantic descriptors, possess the ability to handle such significant intra-class variation and subtle inter-class variation \cite{li2016human}. The complementary attribute and appearance features describe persons with middle-level semantic descriptors and low-level visual properties respectively, thus jointly exploring them could provide more comprehensive and effective representations for pedestrians. Recently, a few of methods \cite{lin2017improving, schumann2017person, shi2015transferring, su2018multi} leverage human attributes to improve image-based person re-identification. These methods utilize the pre-trained attribute recognition models on additional datasets to abstract attribute feature responses on static images, which are in turn utilized to recognize the identities of pedestrians. Nevertheless, these methods model attribute features individually without considering of the semantic contextual correlation among them. Some human attributes usually co-occur, while some ones are impossible to appear simultaneously. Therefore, the status (presence/absence) of a specific attribute could offer crucial information to help recognize the status of other relevant attributes. Besides, an attribute is often corresponding to one latent region of a video frame rather than the whole frame.

In this work, we propose a novel Temporal Attribute-Appearance Learning Network (TALNet) for video-based person re-identification to learn comprehensive and effective pedestrian features. TALNet simultaneously exploits human attributes and appearance by multi-task learning fashion. It explores hard visual attention and temporal-semantic context for attributes, and spatial-temporal dependencies among parts of the body for appearance as well as the complementation between attributes and appearance. As illustrated in Fig.\ref{network}, TALNet is composed of an attribute branch network for learning attribute features, an appearance branch network for learning appearance features, and a base network for extracting low-level visual features. The attribute branch network is composed of a spatial attention block, a temporal-semantic context block as well as several fully connected (FC) layers. The spatial attention block explores the visual attention for each human attribute with only identity-level supervised labels and enforces the network to concern the corresponding patches of video frames. The temporal-semantic context block simultaneously learns the semantic contextual information among attributes within each video frame and the temporal contextual information for each attribute among consecutive frames, towards boosting the performance of attribute recognition. The appearance branch network consists of two Gated Recurrent Unit (GRU) layers \cite{cho2014learning}, four pooling layers as well as several convolution layers and FC layers. The network learns effective appearance representation from both whole body and body parts with spatial-temporal dependencies among them. Moreover, the base network is based on the popular ResNet-50 architecture \cite{resnet} to abstract low-level visual features of pedestrians. On account of the three networks, TALNet is capable of exploiting the complementation of human appearance and attributes, and learning effective video representation, towards accurate re-identification results. We evaluate the proposed TALNet on two video datasets, MARS and iLIDS-VID, and verify it obtains superior performance over all existing re-identification methods. Considering that there is no existing video dataset for person re-identification annotating human attributes, we also make some extra works to annotate ID-level attributes for each pedestrian in these datasets.

In summary, our contributions are the following: (1) We propose a new Temporal Attribute-Appearance Learning Network (TALNet) for video-based person re-identification. (2) We develop an attribute branch network with a spatial attention block and a temporal-semantic context block, by exploiting hard visual attention and latent temporal-semantic context across video frames. (3) We design an appearance branch network with two GRU layers, which learns global and local appearance features from videos with spatial-temporal dependencies among them. (4) We label a set of human attributes for pedestrians in MARS and iLIDS-VID datasets. 

\section{Related works}
Person re-Identification for static images has been extensively studied. Recently, researchers pay attention to video person re-identification. In this section, we briefly review the two categories of related works.

\subsection{Image-based Person Re-Identification}

Early image-based approaches mainly focus on designing discriminative hand-crafted features from static images or learning distance metric for feature matching. Representative hand-crafted features include Local Binary Patterns (LBP) \cite{li2013locally}, Color Names \cite{chen2016similarity} and Scale Invariant Feature Transforms \cite{zhao2013unsupervised}, \textit{etc}. Some more complicated hand-crafted features have also been proposed. For instance,  Matsukawa \textit{et al.} \cite{matsukawa2016hierarchical} presented a hierarchical Gaussian descriptor for person re-identification, which modeled both mean and covariance information of pixel features in each of the patch and region hierarchies. Liao \textit{et al.} \cite{liao2015person} proposed an descriptor called Local Maximal Occurrence (LOMO) feature. It explored the horizontal occurrence of local representations and maximized the occurrence to make a stable representation against viewpoint variations. 
Besides hand-crafted features, similarity/metric learning techniques \cite{zheng2011person, cheng2016person} has been widely applied/designed for person re-identification against light, view and pose changes. For example, Li \textit{et al.} \cite{li2013learning} proposed to learn a decision function for person re-identification. The proposed second-order formulation generalized from traditional metric learning technique by offering a locally adaptive decision rule.


In recent years, deep learning based approaches \cite{hou2019interaction} have achieved prominent progress through learning deep appearance representations. For instance, Chen \textit{et al.} \cite{chen2020salience} proposed Salience-guided Cascaded Suppression Network (SCSN) for person re-identification. It handled the issues of extracting  discriminative features and integrating these features. The designed suppression technique enabled the model to adaptively mine potential significant information on different important levels. In addition, a few of recent methods \cite{schumann2017person,shi2015transferring,su2018multi,lin2017improving} have attempted to use human attribute to improve the capacity of feature representation. The results from these methods have verified the robustness of attribute features against aforementioned challenges. For instance, Shi \textit{et al.} \cite{shi2015transferring} trained an attribute recognition model on the additional fashion photography dataset, and then used it to abstract visual features for person matching. Lin \textit{et al.} \cite{lin2017improving} labeled human attributes for pedestrians on the two widely-used datasets, \textit{i.e.}, Market-1501 and DukeMTMC-reID. Furthermore, they designed an attribute-person recognition (APR) network to simultaneously learn the appearance and attribute representations from images for pedestrian matching. Su \textit{et al.} \cite{su2018multi} proposed a Multi-Task Learning with Low Rank Attribute Embedding (MTL-LORAE) for person re-identification. The framework integrated low-level visual features with attribute features as the descriptors for pedestrians. Moreover, it introduced a low-rank attribute embedding to improve the accuracy of such descriptions.

\begin{figure*}[!t]
\centering
\includegraphics[width=1.0\textwidth]{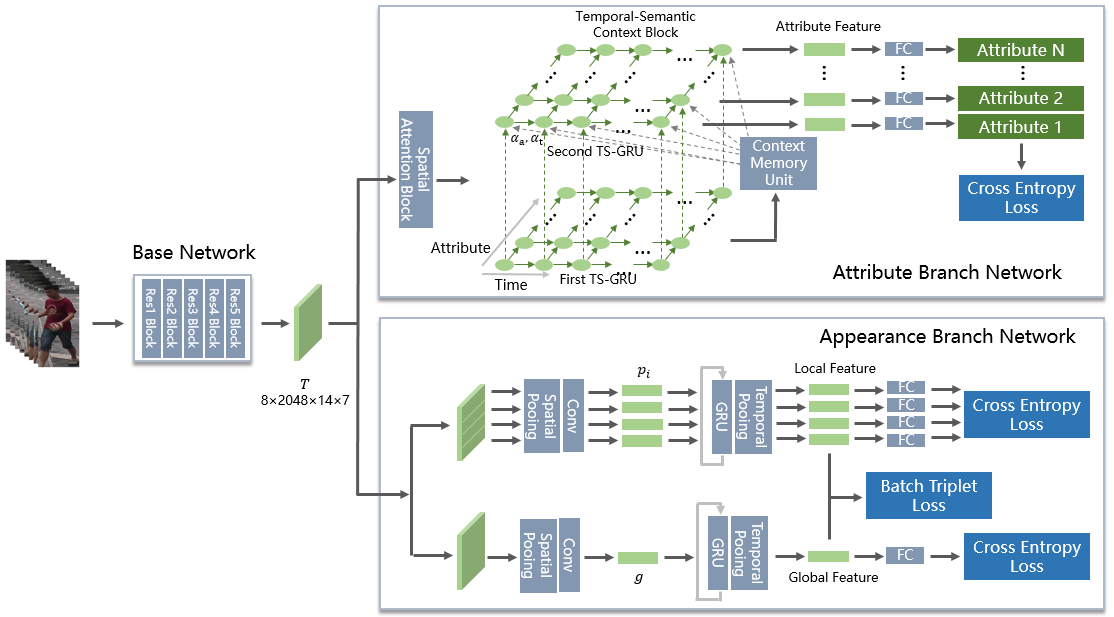}
\caption{The whole architecture of TALNet is composed of a base network for extracting low-level feature maps, an attribute branch network for learning attribute features and an appearance branch network for learning appearance representations.}
\label{network}
\end{figure*}

\subsection{Video-based Person Re-Identification}
Traditional approaches \cite{liu2015spatio} for video-based person re-identification tend to employ hand-crafted video-level descriptors or learn appropriate distance metric. For example, Wang \textit{et al.} \cite{wang2014person} formulated a method of fusing HOG3D features \cite{klaser2008spatio} and optical flow energy profile to obtain a video-level  pedestrian representation. Tao \textit{et al.} \cite{tao2013person} presented regularized smoothing KISS (RE-KISS) metric learning, which seamlessly integrated smoothing and regularization techniques for robustly estimating covariance matrices. It both enlarged the under-estimated small eigenvalues and suppressed the over-estimated large eigenvalues of the estimated covariance matrix.


Several deep learning based approaches \cite{mclaughlin2016recurrent, zhou2017see, xu2017jointly, liu2019spatial, li2019global} for video-based person re-identification have been proposed and shown superior performance compared to traditional approaches. McLaughlin \textit{et al.} \cite{mclaughlin2016recurrent} presented a typical siamese network to capture appearance and motion features from pedestrian videos with optical flow information. It then utilized a recurrent layer and a temporal pooling layer to generate video-level representation. Xu \textit{et al.} \cite{xu2017jointly} presented a joint Spatial and Temporal Attention Pooling Network (ASTPN). It extended the standard CNN-RNN by using a spatial-pooling on feature map from CNN and an attentive temporal-pooling from the output of RNN.  Li \textit{et al.} \cite{li2018diversity} presented a spatio-temporal attention mechanism and a diversity regularization. The spatio-temporal attention was used to localize discriminative image regions and aggregate the learned local feature across time. The diversity regularization ensured the spatial attention did not concentrate on the same body part. Li \textit{et al.} \cite{li2019multi} proposed a two-stream convolution network to simultaneously abstract spatial-temporal visual cues. A temporal stream is developed by inserting several Multi-scale 3D convolution layers into a 2D CNN network. The spatial stream is designed with a 2D CNN for spatial feature extraction. Subramaniam \textit{et al.} \cite{Subramaniam_2019_ICCV} proposed a Co-segmentation inspired video re-identification deep architecture and formulated a Co-segmentation based Attention Module (COSAM) that activates a common set of salient features across multiple frames of a video via mutual consensus in an unsupervised manner.


\section{The proposed method}

In this section, we first describe the overall architecture of TALNet. Then, the main components of TALNet are presented.

\subsection{Overall Architecture}
The person re-identification dataset is denoted as $X=\{\bm{x}_i\}_{i=1}^V$, which contains $V$ video sequences from $P$ pedestrians collected by non-overlapping cameras with the person identities $Y=\{\bm{y}_i\}_{i=1}^V$. The key for person re-identification is to learn discriminative and robust descriptors of pedestrians from video sequences. We propose a novel Temporal Attribute-Appearance Learning Network (TALNet), which simultaneously explores the complementation between human attributes and appearance, hard visual attention and temporal-semantic context for attributes, and spatial-temporal dependencies among body parts for appearance. As illustrated in Fig.\ref{network}, TALNet is composed of an attribute branch network for modeling human attributes, an appearance branch network for characterizing visual appearance and a base network for extracting low-level visual details. It firstly divides each video sequence into several consecutive non-overlap clips $C=\{\bm{c}_i\}_{i=1}^L$ (each clip contains $T$ frames), then takes each clip as input and extracts clip-level pedestrian features from them. These clip-level pedestrian features are applied with average pooling strategy to form the final video-level pedestrian feature.

The base network is based on the popular ResNet-50 model\cite{resnet}, which contains five residual layers. Each residual layer contains several bottleneck building blocks, and each block is composed of three convolution layers with Batch Normalization (BN) layer\cite{ioffe2015batch} , Rectified Linear Units (ReLU) layer \cite{nair2010rectified} and max pooling layer. We develop an attribute branch network with a spatial attention block and a temporal-semantic context block to effectively abstract attribute representation $f(\bm{x}_i)_{att}$. The spatial attention block enforces the network to concentrate on latent local regions corresponding to each attribute for each video frame. The temporal-semantic context block consists of two temporal-semantic GRU layers (TS-GRUs) and a context memory unit, which employs the semantic correlation among attributes within each video frame and temporal correlation for each attribute across video frames simultaneously, towards obtaining precise and effective attribute representations. Moreover, we design an appearance branch network to abstract appearance representation $f(\bm{x}_i)_{app}$ from both whole body and body parts of pedestrians. The full-body feature describe the global visual appearance, while the body-part features describes fine-grained local visual details. Two GRU layers within the appearance branch network are utilized to exploit the temporal information among consecutive frames, and encode the global and local frame-level appearance features sequentially. These two representations are then aggregated into the clip-level appearance representation by temporal pooling operation. The appearance and attribute features are optimized for the tasks of person re-identification and attribute recognition respectively, by multi-task learning fashion. During testing stage, the learned two features are concatenated to generate the effective pedestrian representation. The similarity between two video sequences $\bm{x}_i$ and $\bm{x}_j$ is calculated:
\begin{equation}
\begin{aligned}
& S(\bm{x}_i,\bm{x}_j) \\= &{\parallel{[f(\bm{x}_i)_{app},\lambda_0 f(\bm{x}_i)_{att}] -[f(\bm{x}_j)_{app},\lambda_0 f(\bm{x}_j)_{att}]}\parallel}_2^2
\label{Eq1}
\end{aligned}
\end{equation}
where $\lambda_0$ denotes the balance weight between the appearance and attribute features.

\subsection{Attribute Branch Network}
An attribute branch network is developed to generate robust middle-level semantic descriptors of pedestrians from video sequences via the task of attribute recognition. An attribute usually arises from different local regions within video frames. Consequently, the network is expected to focus on the corresponding regions when modeling attributes. Nevertheless, such regions for attributes are not localized with ground-truth. Different attributes of a pedestrian in each video frame are semantically related. The status of presence or absence for a specific attribute is often beneficial for inferring the accurate status of other associated attributes. For instance, human attributes \textit{short hair} and \textit{wearing trousers} are likely to co-occur, while the attributes \textit{male} and \textit{wearing a dress} may mutually exclusive. The exploration of such semantic context among attributes could improve the accuracy of attribute recognition. Moreover, temporal context information across video frames could also boost the learning of attribute representation.

Considering that, we design a novel attribute branch network with a spatial attention block and a temporal-semantic context block. The proposed spatial attention block contains $2N+1$ convolution layers and $2N$ FC layers ($N$ denotes the amount of pedestrians' attributes), driving the network to concentrate on the potentially attribute-relevant region within video frames, and obtain the initial attribute features. The temporal-semantic context block contains two TS-GRUs and a context memory unit, which simultaneously captures the semantic context among attributes within each video frame and the temporal context for each attribute across video frames, thus boosts the learning of attribute representations. The first TS-GRU takes the initial attribute features in sequence along temporal and attribute dimensions, towards generating the primitive hidden representations, which represent the relative importance among different frames and different attributes in each frame, respectively. Subsequently, a context memory unit is developed to generate the temporal attention weights and semantic attention weights according to the primitive hidden representations. Afterwards, the second TS-GRU generates more precise and effective attribute representation on the basis of the primitive hidden representation with the temporal and semantic attention weights.

\begin{figure}[!t]
	\centering
	\includegraphics[width=0.5\textwidth]{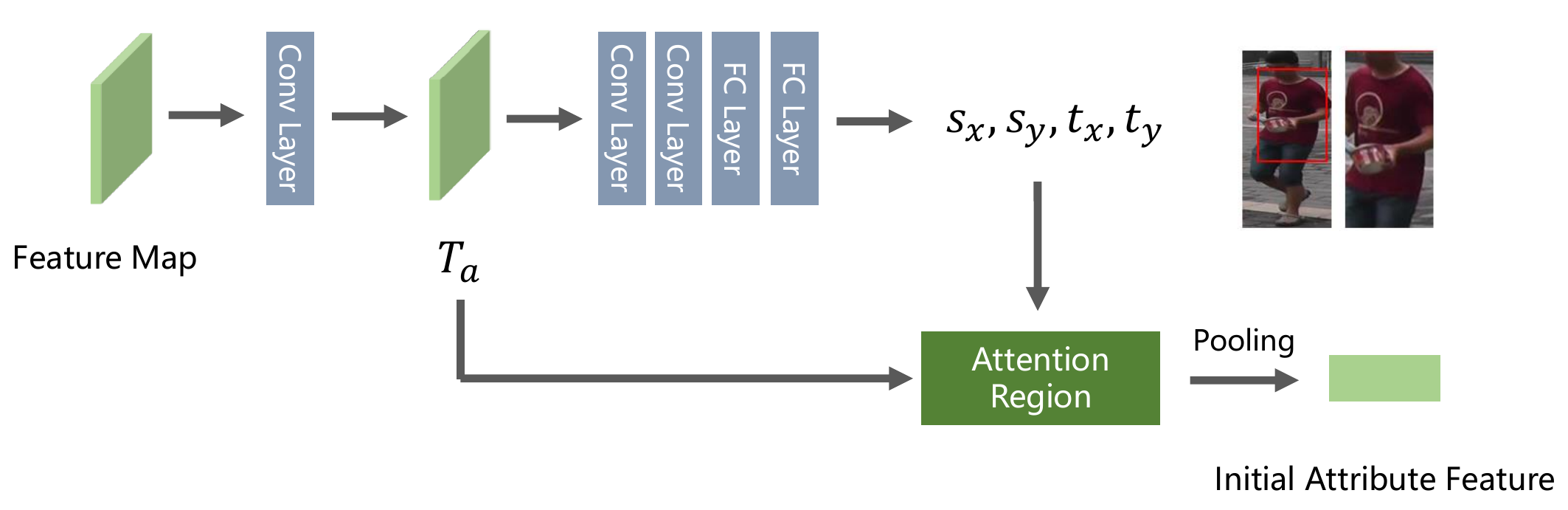}
	\caption{The detailed architecture of the spatial attention block in the attribute branch network.} \label{SA-module}
\end{figure}
\begin{figure}[!t]
	\centering
	\includegraphics[width=0.5\textwidth]{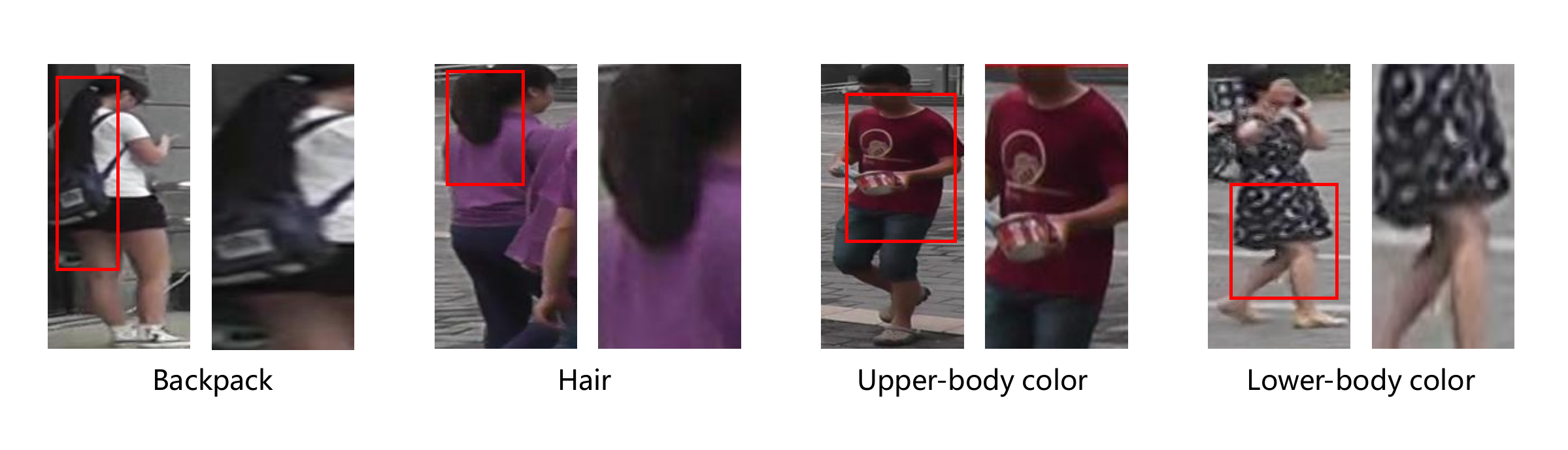}
	\caption{Visualization of the local regions corresponding to certain attributes in some video frames of pedestrians. The attribute regions are highlighted red and cropped.}
	\label{attention}
\end{figure}

As illustrated in Fig.\ref{SA-module}, the spatial attention block contains $2N+1$ convolution layers and $2N$ FC layers. The first convolution layer receives the feature map $\bm{T}$ extracted from the base network as input, and generates a primitive feature $\bm{T}_p$. The convolution kernels of the first convolution layer is $1\times1\times64$, followed by a BN layer and a ReLU layer. For each attribute, two convolution layers and two FC layers take the primitive feature $\bm{T}_p$ as input and obtain the attention region $A_n$ of the attribute. The convolution kernels of these two convolution layers are $5\times5\times32$ and $5\times5\times16$. The channel numbers of these two FC layers are 32 and 4. The first three layers is performed with ReLU operation. The last FC layer outputs a 4-dimension feature vector, which denotes the value of four parameters, $s_x$, $s_y$, $t_x$ and $t_y$. We compute the coordinates of the corresponding attention regions of attributes within each video frame as following:

\begin{equation}
\left(\
\begin{matrix}
x_s \\ y_s
\end{matrix}\
\right)
= 
\left[
\begin{matrix}
s_x & 0 & t_x \\
0 & s_y & t_y
\end{matrix}
\right]
\left(\
\begin{matrix}
x_i \\ y_i \\ 1
\end{matrix}\
\right)
\end{equation}
in which $(x_i,y_i) = \{(0,0),(H,0),(0,W),(H,W)\}$, $H$ and $W$ represent the height and width of video frames. The output $\{(x_{s,j},y_{s,j})\}_{j=1}^4$ are the locations of four vertexes of the attention regions $\{A_i\}_{i=1}^N$. We extract visual cues from the attention region of $\bm{T}_p$ to obtain the initial attribute feature vector $\bm{v}$ for each attribute. Fig.\ref{attention} illustrates the learned attention regions corresponding to certain attributes, such as \textit{backpack}, \textit{hair}, \textit{colors of upper-body and lower-body clothing} within some video frames. From Fig.\ref{attention}, we can observe that the spatial attention block guarantees the attribute branch network to focus on the latent regions related to the attributes and thus improves the accuracy of attribute recognition.

\begin{figure}
	\centering
	\includegraphics[width=0.4\textwidth]{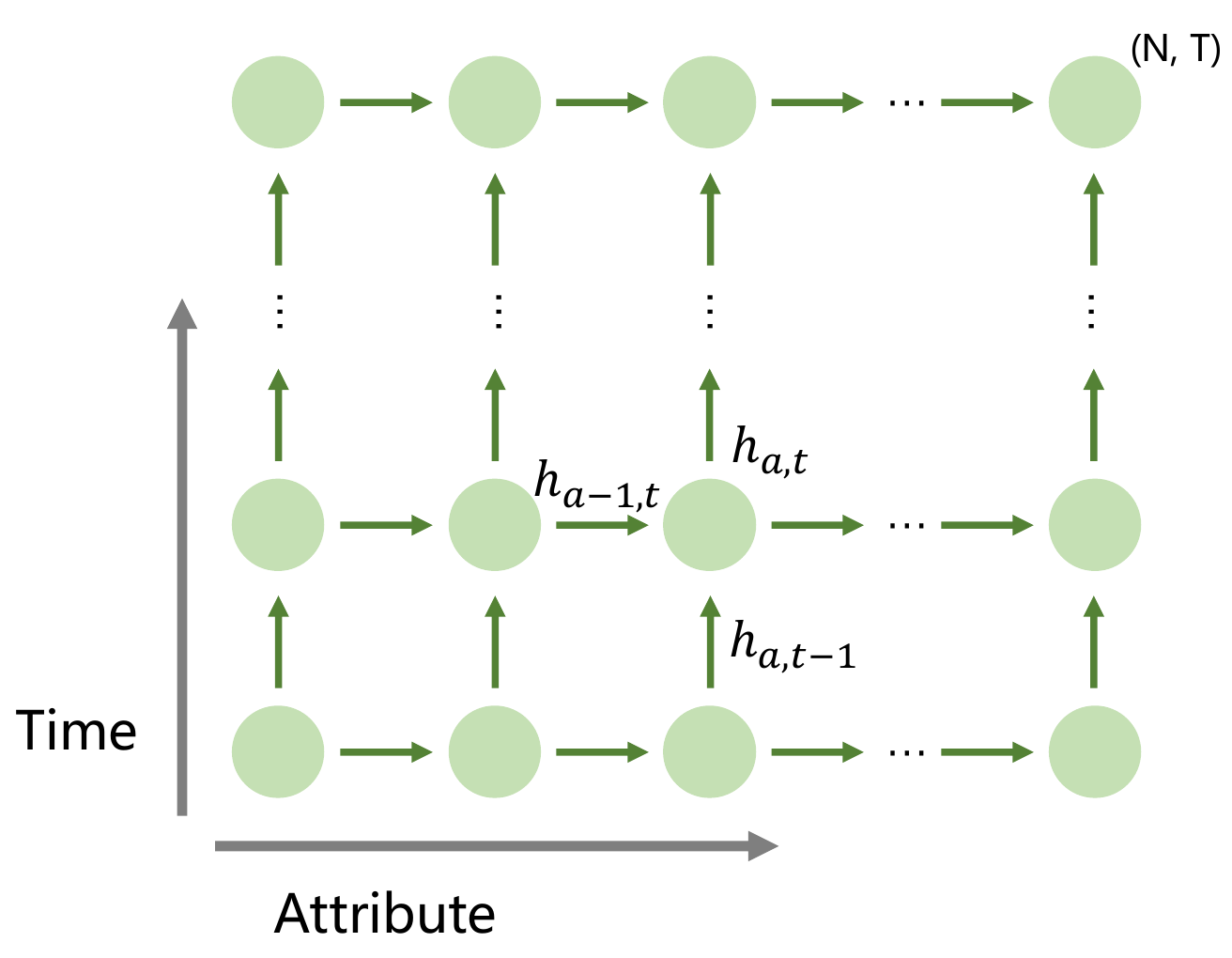}
	\caption{Illustration of the proposed TS-GRU in the attribute branch network. The corresponding initial attribute features in a sequence are fed along the time and attribute directions.} \label{TS-GRU}
\end{figure}

The temporal-semantic context block contains two TS-GRU layers and a context memory unit. As shown in Fig.\ref{TS-GRU}, the initial attribute features of different attributes over different frames are fed into the first TS-GRU along the attribute direction as well as the time direction. At the attribute-time step $(a, t)$, the first TS-GRU takes the initial attribute feature $\bm{v}_{a, t}$, the hidden representation of the previous frame $\bm{h}_{a,t-1}$, the hidden representation of the previous attribute $\bm{h}_{a-1,t}$ as inputs, and generates the primitive hidden representation, where $a \in \{1,\cdots,N\}$ ($N$ denotes the number of human attributes), $t \in \{1,\cdots,T\}$ ($T$ denotes the amount of frames in video clips). The first TS-GRU is formulated as following:
\begin{equation}
\begin{aligned}
\bm{z}_{a,t}^{(A)} &= \sigma (\bm{W}_z^{(A)} [\bm{h}_{a-1,t}, \bm{v}_{a,t}])
\\
\bm{z}_{a,t}^{(T)} &= \sigma (\bm{W}_z^{(T)} [\bm{h}_{a,t-1}, \bm{v}_{a,t}])
\\
\bm{r}_{a,t}^{(A)} &= \sigma (\bm{W}_r^{(A)} [\bm{h}_{a-1,t}, \bm{v}_{a,t}])
\\
\bm{r}_{a,t}^{(T)} &= \sigma (\bm{W}_r^{(T)} [\bm{h}_{a,t-1}, \bm{v}_{a,t}])
\\
\bm{\tilde{h}}_{a,t} &= tanh (\bm{W}_h [\bm{r}_{a,t}^{(A)} \odot \bm{h}_{a-1,t}, \bm{r}_{a,t}^{(T)} \odot \bm{h}_{a,t-1}, \bm{v}_{a,t}])
\\
\bm{h}_{a,t} &= \bm{z}_{a,t}^{(A)} \odot \bm{h}_{a-1,t} + \bm{z}_{a,t}^{(T)} \odot \bm{h}_{a,t-1}
\\& + (1- \bm{z}_{a,t}^{(A)} - \bm{z}_{a,t}^{(T)}) \odot \bm{\tilde{h}}_{a,t}
\end{aligned}\label{EQ-GRU}
\end{equation}
where $\bm{z}^{(A)}$, $\bm{z}^{(T)}$, $\bm{r}^{(A)}$ and $\bm{r}^{(T)}$ denote the attribute update gate, the temporal update gate, the attribute reset gate and the temporal reset gate, respectively. $\bm{W}_z^{(A)}$, $\bm{W}_z^{(T)}$, $\bm{W}_r^{(A)}$ and $\bm{W}_r^{(T)}$ are the parameter matrixes. $\bm{h}_{a,t}$ represents the generated hidden representation at the attribute-time step $(a, t)$.

The context memory unit first averages the primitive hidden representations of all the attributes in each video frame and all frames of each attribute, towards generating the semantic context memory $\bm{F}^{(S)}$ and the temporal context memory $\bm{F}^{(T)}$. $\bm{F}^{(S)}$, $\bm{F}^{(T)}$ include the global semantic knowledge  and temporal contextual information, which are computed as following:
\begin{equation}
\bm{F}_a^{(S)} = \frac{1}{T} \sum_{t=1}^T \bm{h}_{a,t}
\end{equation}
\begin{equation}
\bm{F}_t^{(T)} = \frac{1}{N} \sum_{a=1}^N \bm{h}_{a,t}
\end{equation}
Then it utilizes the two context memories together with the hidden representations $\bm{h}_{a,t}$ to obtain the temporal attention score and the semantic attention score, impelling the network to concentrate on the crucial temporal context among video frames and the informative semantic context among the attributes. The semantic attention score $a^{(S)}$ and temporal attention score $a^{(T)}$ are computed as following:
\begin{equation}
\begin{aligned}
e_{a,t}^{(S)} &= \bm{W}_{a1} \cdot ReLU (\bm{W}_{a2} [\bm{h}_{a,t}, \bm{F}_a^{(S)}])
\\
a_{a,t}^{(S)} &= \frac{exp(e_{a,t}^{(S)})}{\sum_{a=1}^N exp(e_{a,t}^{(S)})}
\end{aligned}
\end{equation}
\begin{equation}
\begin{aligned}
e_{a,t}^{(T)} &= \bm{W}_{t1} \cdot ReLU (\bm{W}_{t2} [\bm{h}_{a,t}, \bm{F}_t^{(T)}])
\\
a_{a,t}^{(T)} &= \frac{exp(e_{a,t}^{(T)})}{\sum_{t=1}^T exp(e_{a,t}^{(T)})}
\end{aligned}
\end{equation}
where $\bm{W}_{a1}$, $\bm{W}_{a2}$, $\bm{W}_{t1}$, $\bm{W}_{t2}$ denote the parameter matrixes. $a_{a,t}^{(A)}$ and $a_{a,t}^{(T)}$ refer to the normalized semantic and temporal attention scores of the input attribute feature at the step $(a,t)$. For the second TS-GRU, the formulation for generating $\bm{z}^{(A)}$, $\bm{z}^{(T)}$, $\bm{r}^{(A)}$, $\bm{r}^{(T)}$ and $\bm{\tilde{h}}_{a,t}$ is similar to Eq.(\ref{EQ-GRU}). Afterwards, the hidden representation $\bm{h}_{a,t}^\prime$ is computed:
\begin{equation}
\begin{aligned}
\bm{h}_{a,t}^\prime &= a_{a,t}^{(A)} \cdot \bm{z}_{a,t}^{(A)} \odot \bm{h}_{a-1,t} + a_{a,t}^{(T)} \cdot \bm{z}_{a,t}^{(T)} \odot \bm{h}_{a,t-1}
\\& + (1-a_{a,t}^{(A)} \cdot \bm{z}_{a,t}^{(A)} - a_{a,t}^{(T)} \cdot \bm{z}_{a,t}^{(T)}) \odot \bm{\tilde{h}}_{a,t}
\end{aligned}
\end{equation}
The hidden representation of the last frame $\bm{h}_{a,T}^\prime$ is the ultimate feature of attribute $a$. Finally, each extracted attribute representation is fed into a $m$-dimension FC layer, which is optimized by cross entropy loss ($m$ denotes that the certain attribute contains $m$ categories). The attribute branch network employs the spatial attention block and the temporal-semantic context block to learn hard visual attention and temporal-semantic contextual information for attributes, which effectively improves the performance of attribute recognition.

\subsection{Appearance Branch Network}
An appearance branch network is designed to simultaneously explore the global and local appearance representations for person re-identification. The full-body features describe the global visual appearance of pedestrians, while the part-body features describe the fine-grained local visual cues within each body part. Moreover, two GRU layers are utilized to capture the temporal contextual information contained in videos, which encode the frame-level full-body and part-body representations sequentially and generate more comprehensive and robust clip-level appearance features.

As illustrated in Fig.\ref{network}, the appearance branch network contains two convolution layers, four pooling layers, two GRU layers and five FC layers.  Each convolution layer is connected with a BN operation and a ReLU operation. The appearance branch network receives the feature map $\bm{T}$ as input, which is generated from the base network. The appearance branch network divides $\bm{T}$ into $H$ horizontal stripes (we set $H=4$ in this work). Subsequently, the feature tensor $\bm{T}$ and $H$ horizontal stripes are fed into a global branch and a local branch to abstract full-body and part-body representations, respectively. The feature vector of activations along the channel dimension is defined as a column vector in a feature map. The two branches in this network adopt spatial pooling layers (meaning pooling operation) to aggregate all column vectors in each horizontal stripe to produce $H$ horizontal part-level column vectors or in the whole feature map $\bm{T}$ to generate a global-level column vector, respectively. Afterwards, the network utilizes two convolution layers to reduce the dimension of part- and global-level column vectors, and generate the final global-level feature vector $\bm{g}$ and $H$ part-level feature vectors $\bm{p}_i$ $(i = 1,2,\cdots,H)$. Then, two GRU layers are applied to explore the temporal information, which sweep the global-level and part-level features over frames sequentially. The global GRU layer is formulated as following:
\begin{equation}
\begin{aligned}
\bm{z}_t &= \sigma (\bm{W}_z [\bm{h}_{t-1}, \bm{g}_t])
\\
\bm{r}_t &= \sigma (\bm{W}_r [\bm{h}_{t-1}, \bm{g}_t])
\\
\bm{\tilde{h}}_{t} &= tanh (\bm{W}_h [\bm{r}_{t} \odot \bm{h}_{t-1}, \bm{g}_{t}])
\\
\bm{h}_{t} &= (1- \bm{z}_{t}) \odot \bm{h}_{t-1} + \bm{z}_{t} \odot \bm{\tilde{h}}_{t}
\end{aligned}\label{GRU}
\end{equation}
in which $\bm{g}_t$ denotes the global-level feature vector of frame $t$. $\bm{r}_t$ and $\bm{z}_t$ refer to reset and update gates. $\bm{W}_z$, $\bm{W}_r$ and $\bm{W}_h$ are the learnable parameter matrixes. $\bm{h}_{t}$ denotes the learned hidden representation at time step $t$. We utilize a temporal pooling layer (meaning pooling operation) to average all the output hidden representation for obtaining the clip-level global feature. Meanwhile, we generate $H$ clip-level part features through the local GRU layer in a similar way. Finally, the clip-level global and part features are taken into the classifier layers which are implemented with FC layers, and optimized by cross entropy loss. In addition, the clip-level global and part features are concatenated together and optimized by triplet loss, following the previous works \cite{zhang2019densely,zheng2019re,tay2019aanet}. During testing stage, the abstracted global and part representations are fused into the appearance representation of pedestrian. The appearance representation is employed for pedestrian matching with the learned attribute feature.


\subsection{Model Optimization}
We employ triplet loss and cross entropy loss to optimize the appearance branch network. Triplet loss is equipped with 
hard mining strategy \cite{hermans2017defense}. We randomly select $I$ pedestrians and $V$ video clips for each pedestrian to form a batch. Consequently, there are $I \times V$ video clips in each batch, totally. For each sample in the batch, the hardest positive sample and hardest negative sample are picked to constitute a triplet, which is employed for calculated the triplet loss $L_{tri}$:

\begin{equation}
\begin{aligned}
L_{tri} (\bm{f}) = \overbrace{\sum_{i=1}^{I} \sum_{j=1}^{V}}^{all\ anchors} &[ m + \overbrace{\max \limits_{m=1 \cdots V} D(\bm{f}_{i,a}, \bm{f}_{i,m})}^{hardest\ positive} \\ 
&- \underbrace{\min \limits_{n=1 \cdots I \atop {m=1 \cdots K \atop n \neq i}} D(\bm{f}_{i,a}, \bm{f}_{n,m})}_{hardest\ negative}]_+
\end{aligned}
\end{equation}
where $\bm{f}$ denotes the appearance features. $D(,)$ refers to the Euclidean distance between two features. Cross entropy loss function is equipped with label smoothing regularization \cite{szegedy2016rethinking}. The formulation of this loss is defined as following:

\begin{equation}
L_{ide} (\bm{q}) = - \frac{1}{I \times V} \sum_{i=1}^{I} \sum_{j=1}^{V} \bm{p}_{i,j} log((1-\epsilon)\bm{q}_{i,j}+\frac{\epsilon}{G})
\label{Eq3}
\end{equation}
where $\bm{p}_{i,j}$ denotes the ground truth pedestrian identity, $\bm{q}_{i,j}$ is the computed prediction score for the sample ${i,j}$. $G$ represents the number of pedestrians and $\epsilon$ is the smoothing parameter. Triplet loss and cross entropy loss are simultaneously employed to optimize the global and part features. Thus, the final loss for the appearance branch network is defined as following:

\begin{equation}
\begin{aligned}
L_{app} & = L_{tri} + L_{ide}
\\& = L_{tri}(\bm{g}) + \sum _{h=1}^H L_{tri}(\bm{p}_h) + L_{ide}(\bm{q}_g) + \sum _{h=1}^H L_{ide}(\bm{q}_{p,h})
\end{aligned}
\end{equation}
where $\bm{g}$ refers to the global feature and $\bm{p}_h$ refers to the $h$-th part feature. $\bm{q}_g$ and $\bm{q}_{p,h}$ represent the prediction scores of the global feature and $h$-th part feature, respectively.

In addition, we adopt cross entropy loss for attribute recognition, which is defined as following:

\begin{equation}
L_{att} = \sum_{n=1}^{N} L_{ide}(\bm{q}_{A,n})
\end{equation}
where $\bm{q}_{A,n}$ denotes the computed prediction scores of the $n$-th attribute, $N$ denotes the number of attributes. By jointing optimizing the tasks of person re-identification and attribute recognition, TALNet is able to predict pedestrian identities and attributes labels. The total loss for TALNet is shown as following:

\begin{equation}
L = L_{app} + \lambda \cdot L_{att}
\label{Eq2}
\end{equation}
in which $\lambda$ is the balance parameter.

\begin{figure}
\centering
\includegraphics[width=0.5\textwidth]{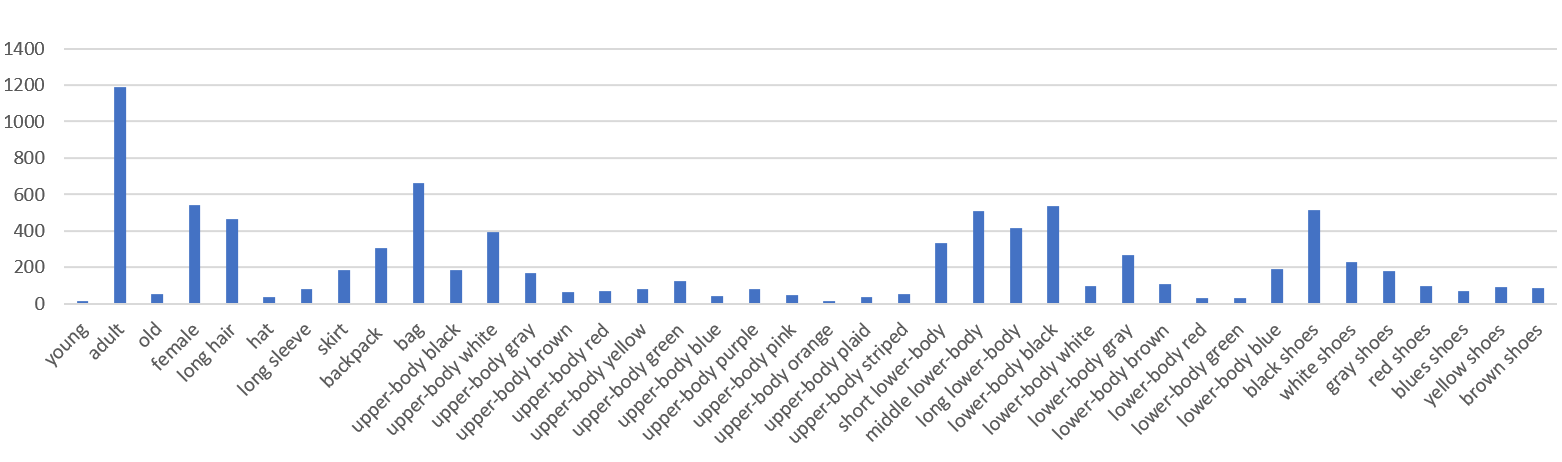}
\caption{The distribution of attributes on MARS dataset. For each attribute, we show the number of pedestrians containing this attribute.} \label{MARSAtt}
\end{figure}

\begin{figure}
\centering
\includegraphics[width=0.5\textwidth]{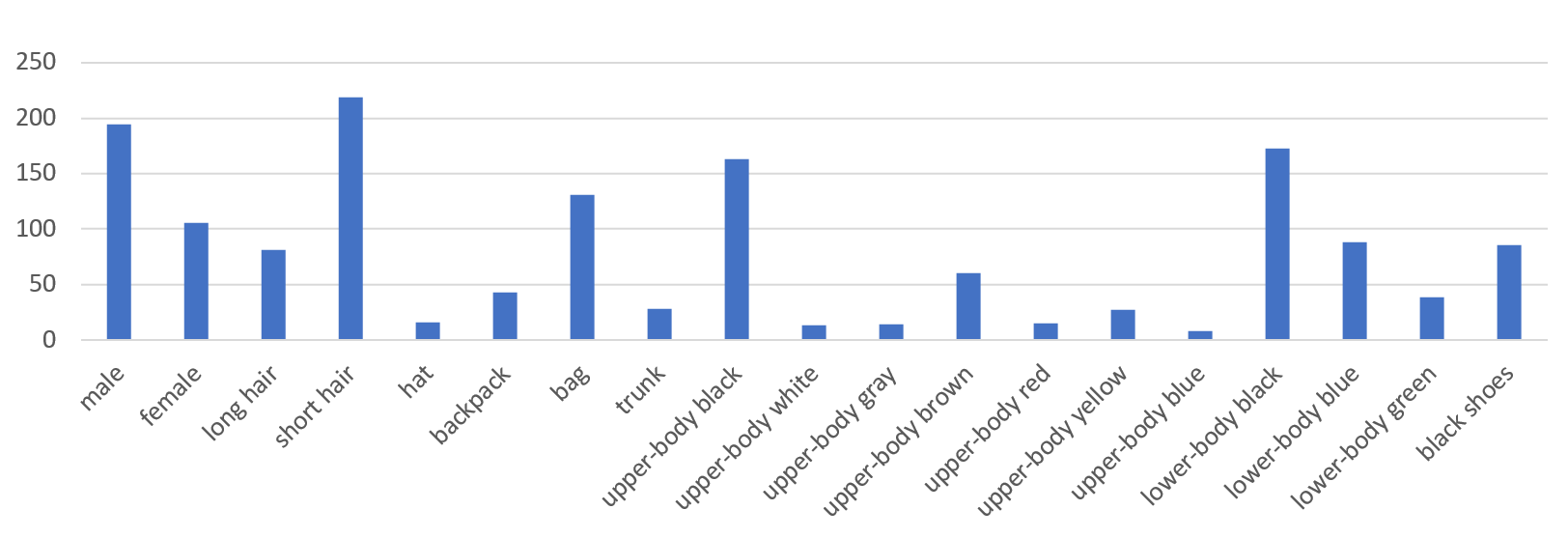}
\caption{The distribution of attributes on iLIDS-VID dataset. For each attribute, we show the number of pedestrians containing this attribute.} \label{iLIDSAtt}
\end{figure}

\section{Attribute Annotation}
We manually annotate ID-level attribute labels for pedestrians in MARS \cite{zheng2016mars} and iLIDS-VID \cite{wang2014person} person re-identification datasets. ID-level human attributes represent the information related to the pedestrians themselves, such as \textit{gender}, \textit{age}, \textit{accessory} and \textit{clothing style}. Instance-level human attributes in contrast, describe the information of surrounding scene or an action state lasting for a short time, \textit{e.g.}, \textit{trees}, \textit{riding on the road} and \textit{listing music}. Therefore, ID-level human attributes have more positive effect on boosting the re-identification performance. MARS and iLIDS-VID datasets are collected in different scenarios (university campus or airport arrival hall) and seasons (summer or winter), thus the pedestrians wear distinct clothes. For example, many pedestrians wear short sleeves or dresses in MARS dataset, but most of the pedestrians wear coats and pants in iLIDS-VID dataset. Considering that, we label two different sets of attributes for these two video datasets.

For MARS dataset, we have labeled 14 ID-level attributes: 11 colors of upper-body clothing (red, black, white, blue, brown, green, gray, orange, yellow, purple, pink),  7 colors of lower-body clothing (red, black, white, blue, brown, green, gray), plaid pattern on upper-body clothing (no, yes), striped pattern on upper-body clothing (no, yes), length of lower-body clothing (short, middle, long), type of lower-body clothing (dress, pants),  7 colors of shoes (red, black, white, blue, gray, brown, yellow), sleeve length (short, long), hair length (short, long), wearing hat (no, yes), carrying backpack (no, yes), carrying bag (no, yes), age (young, adult, old) and gender (female, male). The attribute distribution on MARS dataset is illustrated in Fig.\ref{MARSAtt}.

For iLIDS-VID dataset, we have labeled 9 ID-level attributes: 7 colors of upper-body clothing (red, black, white, blue, brown, gray, yellow), 3 colors of lower-body clothing (black, blue, others), 2 colors of shoes (black, others), hair length (short, long), wearing hat (no, yes), carrying backpack (no, yes), carrying bag (no, yes), carrying trunk (no, yes) and gender (female, male). The attribute distribution on iLIDS-VID dataset is shown in Fig.\ref{iLIDSAtt}.

\section{Experiments}
To evaluate the effectiveness of our proposed TALNet, we conduct several experiments on two widely-used video re-identification datasets. These experiments consist of comparison analysis with state-of-the-art methods and ablation studies in person re-identification task, which run on the server with 4 NIVIDA GTX 1080ti GPUs. The specific experimental settings and results are explained in this section, where bold values highlight the best performance.

\subsection{Experimental Settings}

\subsubsection{Datasets}
The realistic MARS \cite{zheng2016mars} and iLIDS-VID \cite{wang2014person} video datasets are selected for experimental evaluation. MARS dataset is a video-based person re-identification dataset, consisting of 1,261 identities and a total of 20,715 video sequences. Video sequences in MARS dataset are collected automatically by Deformable Part Model detector \cite{felzenszwalb2010object} and Generalized Maximum Multi Clique Problem tracker \cite{dehghan2015gmmcp}. In MARS dataset, 3,248 distractor video sequences are challenging due to poor quality detection or tracking, which significantly increase the difficulty in pedestrians matching. Each identity contains 13.2 video sequences on average, and the length of each video sequence varies from 2 to 920 frames, with an average number of 59.5. Each query has an average number of 3.7 cross-camera ground truths, which contains an average number of 4.2 video sequences captured by the same camera. Following the setting in the work\cite{zheng2016mars}, we fixedly divide this dataset into 625 pedestrians for training and remain 636 pedestrians for testing. iLIDS-VID dataset is a subset of the i-LIDS Multiple-Camera Tracking Scenario dataset \cite{wang2016person}, which was collected at an airport arrival hall under multi-camera topology. iLIDS-VID dataset is very challenging due to different persons with similar clothing, low resolution and occlusions \textit{etc}. It contains 300 identities from 600 video sequences. Each identity captured from two cameras has a pair of video sequences. Each video sequence contains variable length ranging from 23 to 192 video frames, with an average number of 73. Following the work \cite{wang2014person}, this dataset is randomly divided into two parts for training and testing. Each part contains 150 pedestrians.

Considering that the TS-GRU layers in TALNet process human attributes sequentially over video frames, we have to arrange the order of attributes. Nevertheless, they have no fixed and conventional order in reality. A promising way is to employ multi-orders of attributes (\textit{e.g.}, frequent first, random order, or rare first) and fuse their results for obtaining the optimal performance \cite{wang2017attribute}. In following experiments, we exploit two distinct types of orders, \textit{i.e.}, top-down order according to human skeleton structure and fine-abstract order based on the semantic granularity from abstract attributes to fine-grained attributes on these two datasets.

\subsubsection{Evaluation Metrics}
Cumulative Matching Characteristic (CMC) is widely used to quantitatively evaluate the performance of person re-identification algorithms. In CMC curve, the ordinate value corresponding to the abscissa $k$ is recorded as Rank-$k$. It indicates the probability that a query person is truth matched in the Rank-$k$ position. Another important metric is the Mean Average Precision (mAP), which is exploited to evaluate the approaches in a multi-shot setting for person re-identification.

%

\subsubsection{Implementation Details}
The proposed TALNet is implemented with the PyTorch architecture. The initialization of base network adopt the ResNet-50 parameters obtained by pre-training on ImageNet dataset \cite{krizhevsky2012imagenet}. During the training process, the learning rate $lr$ is set to 0.01, the weight deacy and the Nesterov momentum are empirically set to $5e^{-4}$ and 0.9, respectively. Parameter update adopts stochastic gradient descent (SGD) algorithm. The parameters $\lambda_0$ and $\lambda$ in Eq.(\ref{Eq1}) and Eq.(\ref{Eq2}) are set to 0.3, respectively. The parameter $\epsilon$ in Eq.(\ref{Eq3}) is set to 0.1. All inputs are firstly reshaped to the size of $3\times224\times112$ and then normalised with 1.0/256. In addition, the training set is augmented by random erasing probability of 0.3 \cite{zhong2017random}. For triplet loss, 8 identities and 4 clips for each identity are taken as a mini-batch. Each clip contains 8 consecutive frames. TALNet is trained for 400 epochs in total. After 300 epochs, the learning rates $lr$ decay to $0.1\cdot lr$. The whole training process contains two stages. 
During the first stage, we train the base network followed with the appearance branch network until it convergences on person re-identification task, which drives the model to extract helpful feature maps prepared for attribute recognition. During the second stage, we optimize the complete TALNet for attribute recognition and person re-identification simultaneously, towards guiding the model to learn discriminative appearance and attribute representations.

\subsection{Comparison to State-of-the-Art Methods}

\begin{table}[htb]
	\centering
	\newcommand{\tabincell}[2]{\begin{tabular}{@{}#1@{}}#2\end{tabular}}
	\renewcommand{\arraystretch}{1.5}
	\caption{Performance comparison to the related works on MARS dataset.}\label{MARS-state-of-art}
	\begin{tabular}{|c|c|c|c|c|}
		\hline
		{\textbf{Methods}}&\textbf{Rank-1}&\textbf{Rank-5}&\textbf{Rank-20}&\textbf{mAP}\\
		\cline{2-5}
		\hline
		{HistLBP\cite{xiong2014person}+XQDA\cite{liao2015person}}&{18.6}&{33.0}&{45.9}&{8.0}\\
		\hline
		{BoW+KissMe\cite{koestinger2012large}}&{30.6}&{46.2}&{59.2}&{15.5}\\
		\hline
		{IDE+KissMe\cite{zheng2017discriminatively}}&{65.0}&{81.1}&{88.9}&{45.6}\\
		\hline
		{PartsNet\cite{li2017learning}}&{71.8}&{86.6}&{93.1}&{56.1}\\
		\hline
		{QAN\cite{liu2017quality}}&{73.7}&{84.9}&{91.6}&{51.7}\\
		\hline
		{Triplet\cite{hermans2017defense}}&{79.8}&{91.4}&{-}&{67.7}\\
		\hline
		\hline
		{ASPTN\cite{xu2017jointly}}&{44.0}&{70.0}&{81.0}&{-}\\
		\hline
		{end-to-end AMOC\cite{liu2018video}}&{68.3}&{81.4}&{88.9}&{52.9}\\
		\hline
		{JST-RNN\cite{zhou2017see}}&{70.6}&{90.0}&{97.6}&{50.7}\\
		\hline
		{D3DNet\cite{liu2018dense}}&{76.0}&{87.2}&{94.1}&{71.4}\\
		\hline
		{Spatialtemporal\cite{li2018diversity}}&{82.3}&{-}&{-}&{65.8}\\
		\hline
		{Rev-Att\cite{gao2018revisiting}}&{83.3}&{93.8}&{97.4}&{76.7}\\
		\hline
		{3D-CNNA\cite{liao2018video}}&{84.3}&{94.6}&{96.2}&{77.0}\\
		\hline
		RRU\cite{liu2019spatial}&{84.4}&{93.2}&{96.3}&{72.7}\\
		\hline
		M3D\cite{li2019multi}&{84.4}&{93.8}&{97.7}&{74.1}\\
		\hline
		COSAM\cite{Subramaniam_2019_ICCV}&{84.9}&{95.5}&{97.9}&{79.9}\\
		\hline
		Snippet\cite{chen2018video}&{86.3}&{94.7}&{98.2}&{76.1}\\
		\hline
		STA\cite{fu2019sta}&{86.3}&{95.7}&{97.1}&{80.8}\\
		\hline
		{TALNet}&\textbf{89.1}&\textbf{96.1}&\textbf{98.5}&\textbf{82.3}\\
		\hline
	\end{tabular}
\end{table}

\begin{table}[htb]
	\centering
	\newcommand{\tabincell}[2]{\begin{tabular}{@{}#1@{}}#2\end{tabular}}
	\renewcommand{\arraystretch}{1.5}
	\caption{Performance comparison to the related works on iLIDS-VID dataset.}\label{iLIDS-VID-state-of-art}
	\begin{tabular}{|c|c|c|c|c|}
		\hline
		{\textbf{Methods}}&\textbf{Rank-1}&\textbf{Rank-5}&\textbf{Rank-10}&\textbf{Rank-20}\\
		\cline{2-5}
		\hline
		{DTIVD\cite{karanam2015person}}&{25.9}&{48.2}&{57.3}&{68.9}\\
		\hline
		{DS-ranking\cite{wang2016person}}&{39.5}&{61.6}&{71.7}&{81.0}\\
		\hline
		{STAP-Rep\cite{liu2015spatio}}&{44.3}&{71.7}&{83.7}&{91.7}\\
		\hline
		{JST-RNN\cite{zhou2017see}}&{55.2}&{86.5}&{-}&{97.0}\\
		\hline
		{T-push\cite{you2016top}}&{56.3}&{87.6}&{95.6}&{98.3}\\
		\hline
		{RNN\cite{mclaughlin2016recurrent}}&{58.0}&{84.0}&{91.0}&{96.0}\\
		\hline
		{ASPTN\cite{xu2017jointly}}&{62.0}&{86.0}&{94.0}&{98.0}\\
		\hline
		{D3DNet\cite{liu2018dense}}&{65.4}&{87.9}&{92.7}&{98.3}\\
		\hline
		{end-to-end AMOC\cite{liu2018video}}&{68.7}&{94.3}&{98.3}&{99.3}\\
		\hline
		{Spatialtemporal\cite{li2018diversity}}&{80.2}&{-}&{-}&{-}\\
		\hline
		{3D-CNNA\cite{liao2018video}}&{81.3}&{-}&{-}&{-}\\
		\hline
		RRU\cite{liu2019spatial}&{84.3}&{96.8}&{-}&{99.5}\\
		\hline
		Snippet\cite{chen2018video}&{85.4}&{96.7}&{-}&{-}\\
		\hline
		{TALNet}&\textbf{87.1}&\textbf{98.2}&\textbf{99.1}&\textbf{99.9}\\
		\hline
	\end{tabular}
\end{table}

\subsubsection{MARS dataset}
\begin{figure}
	\centering
	\includegraphics[width=0.5\textwidth]{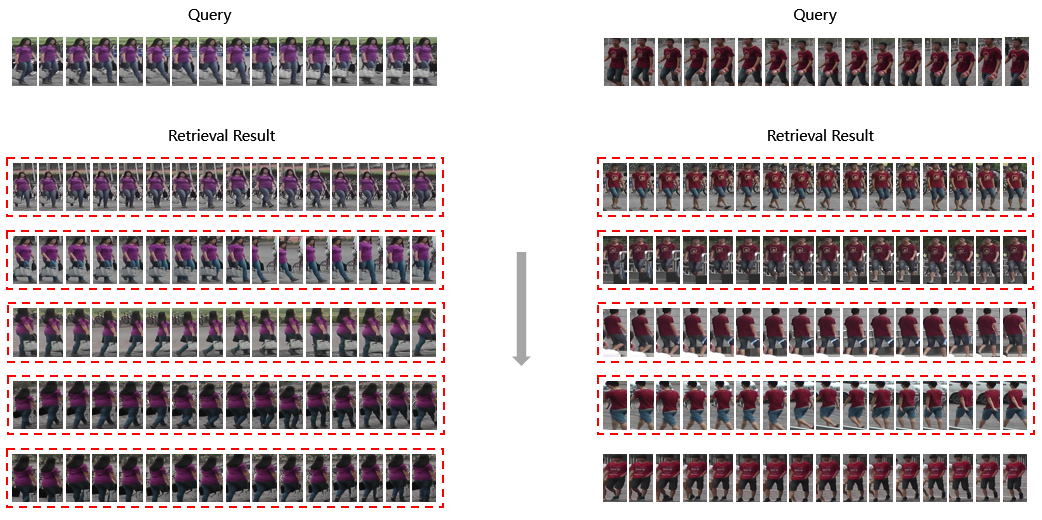}
	\caption{Example retrieval results of the proposed TALNet on the MARS dataset. For each query, the top-5 retrieval results from the gallery are shown from top to down. Correct matches are highlighted red.} \label{visual_MARS}
\end{figure}

In Table \ref{MARS-state-of-art}, these approaches used for comparison include HistLBP \cite{xiong2014person}+XQDA \cite{liao2015person}, BoW+KissMe \cite{koestinger2012large}, IDE+KissMe \cite{zheng2017discriminatively}, PartsNet \cite{li2017learning}, QAN \cite{liu2017quality}, Triplet \cite{hermans2017defense}, ASPTN \cite{xu2017jointly}, end-to-end AMOC \cite{liu2018video}, JST-RNN \cite{zhou2017see}, D3DNet \cite{liu2018dense}, Spatialtemporal \cite{li2018diversity}, Rev-Att \cite{gao2018revisiting}, 3D-CNNA \cite{liao2018video}, RRU \cite{liu2019spatial}, M3D \cite{li2019multi}, COSAM \cite{Subramaniam_2019_ICCV}, Snippet \cite{chen2018video}, STA \cite{fu2019sta}. The first six approaches belong to image-based person re-identification, and the remain approaches belong to video-based person re-identification. Some results of Triplet, ASPTN and Spatialtemporal were not evaluated in their published papers, such as Rank-20 of Triplet \cite{hermans2017defense}. They are marked as {-} in Table \ref{MARS-state-of-art}. From the results, the proposed TALNet obtains superior performance in terms of both Rank-$k$ and mAP over 18 state-of-the-art methods, especially for image-based re-identification algorithms. The Rank-1 and mAP of TALNet reach 89.1\% and 82.3\%, respectively. Compared with the 2nd best STA \cite{fu2019sta}, TALNet improves Rank-1 by 2.8\% and mAP by 1.5\%, respectively. These performance improvement of TALNet owing to jointly learning attribute and appearance representations. Some visual results of person re-identification on MARS dataset are shown in Fig.\ref{visual_MARS}, which again illustrates the advantages of the proposed TALNet.

\subsubsection{iLIDS-VID dataset}
\begin{figure}
		\centering
		\includegraphics[width=0.5\textwidth]{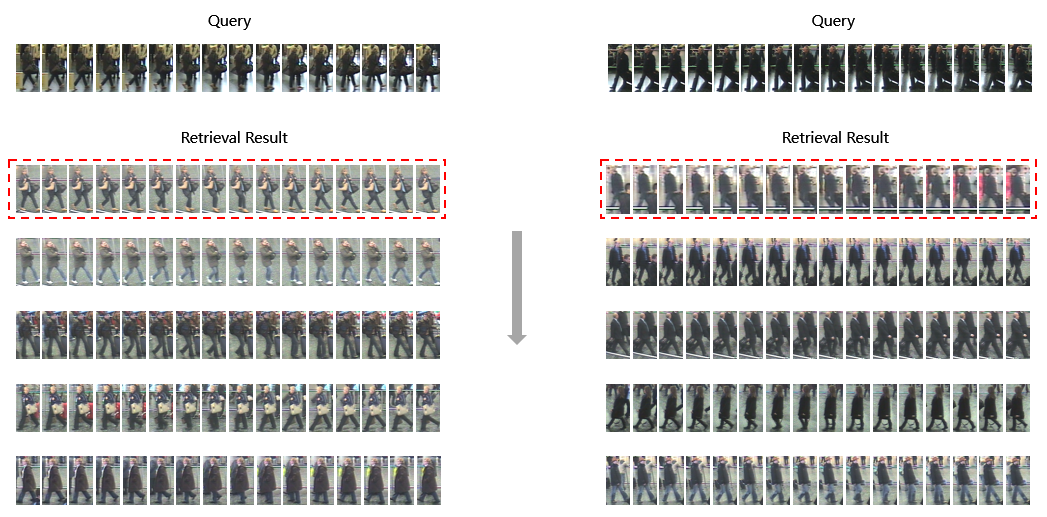}
		\caption{Example retrieval results of the proposed TALNet on the iLIDS-VID dataset. For each query, the top-5 retrieval results from the gallery are displayed from top to down. Correct matches are highlighted red.} \label{visual_iLIDS}
\end{figure}

In Table \ref{iLIDS-VID-state-of-art}, we selected 13 state-of-the-art methods of video-based person re-identification to compare with the proposed TALNet. They include metric learning methods (DTIVD \cite{karanam2015person}, DS-ranking \cite{wang2016person}, T-push \cite{you2016top}), feature learning methods (STAP-Rep \cite{liu2015spatio}), and deep learning methods (JST-RNN \cite{zhou2017see}, RNN \cite{mclaughlin2016recurrent}, ASPTN \cite{xu2017jointly}, D3DNet \cite{liu2018dense}, end-to-end AMOC \cite{liu2018video}, Spatialtemporal \cite{li2018diversity}, 3D-CNNA \cite{liao2018video}, RRU \cite{liu2019spatial} and Snippet \cite{chen2018video}). From Table \ref{iLIDS-VID-state-of-art}, we can observe that the performance of the traditional methods is far inferior to the deep learning methods. TALNet surpasses all the existing methods from Rank-1 to Rank-20. Especially on Rank-1, it boosts the 2nd best compared method Snippet by 1.7\%. The improvement owes to the exploiting of the attribute and appearance branch networks. Fig.\ref{visual_iLIDS} shows some retrieval results by TALNet on iLIDS-VID dataset.

\subsection{Ablation Studies}
To investigate the effectiveness of each component and the design choices including the appearance branch network, the attribute branch network and different pooling strategies, we also perform ablation experiments for TALNet on MARS dataset.
The ablation analysis results of each component within TALNet are summarized in Table \ref{component}. TALNet\_w/o App refers to the base network followed with the attribute branch network, which focuses on extracting attribute feature for pedestrian retrieval. TALNet\_w/o Att represents the base network followed with the appearance branch network, which pays attention to abstracting appearance representation for recognizing pedestrians. As illustrated in Table \ref{component}, TALNet\_w/o App only achieves 65.5\% Rank-1 and 47.9\% mAP, due to the fact that the attribute branch network is not customized for person re-identification. TALNet\_w/o Att obtains 85.9\% Rank-1 and 78.4\% mAP, respectively. Compared to these two models, the integral TALNet yields the best performance of 89.1\% Rank-1 and 82.3\% mAP. The comparison indicates jointly exploring visual appearance and human attributes helps to achieve comprehensive and robust representations, which can significantly improve the accuracy of person re-identification.

\begin{table}[!t]
	\centering
	\newcommand{\tabincell}[2]{\begin{tabular}{@{}#1@{}}#2\end{tabular}}
	\renewcommand{\arraystretch}{1.5}
	\caption{Ablation studies on each component within TALNet on MARS dataset.}\label{component}
	\begin{tabular}{|c|c|c|}
		\hline
		{\textbf{Model}}&\textbf{Rank-1}&\textbf{mAP}\\
		\cline{2-3}
		\hline
		{TALNet\_w/o App}&{65.5}&{47.9}\\
		\hline
		{TALNet\_w/o Att}&{85.9}&{78.4}\\
		\hline
		{TALNet}&{89.1}&{82.3}\\
		\hline
	\end{tabular}
\end{table}

\begin{table}[!t]
	\centering
	\newcommand{\tabincell}[2]{\begin{tabular}{@{}#1@{}}#2\end{tabular}}
	\renewcommand{\arraystretch}{1.5}
	\caption{Ablation studies on each component within the appearance branch network on MARS dataset.}\label{AppNet}
	\begin{tabular}{|c|c|c|}
		\hline
		{\textbf{Model}}&\textbf{Rank-1}&\textbf{mAP}\\
		\cline{2-3}
		\hline
		{AppNet\_w/o global}&{83.5}&{73.5}\\
		\hline
		{AppNet\_w/o local}&{84.2}&{75.0}\\
		\hline
		{AppNet\_w/o GRU}&{84.5}&{75.4}\\
		\hline
		{AppNet}&{85.9}&{78.4}\\
		\hline
	\end{tabular}
\end{table}

\begin{table}[!t]
	\centering
	\newcommand{\tabincell}[2]{\begin{tabular}{@{}#1@{}}#2\end{tabular}}
	\renewcommand{\arraystretch}{1.5}
	\caption{Ablation studies on each component within the attribute branch network on MARS dataset.}\label{AttNet}
	\begin{tabular}{|c|c|c|}
		\hline
		{\textbf{Model}}&\textbf{Rank-1}&\textbf{mAP}\\
		\cline{2-3}
		\hline
		{AttNet\_w/o ST}&{54.5}&{37.9}\\
		\hline
		{AttNet\_w/o T}&{60.2}&{40.6}\\
		\hline
		{AttNet\_w/o C}&{62.7}&{45.8}\\
		\hline
		{AttNet}&{65.5}&{47.9}\\
		\hline
	\end{tabular}
\end{table}

\begin{table}[!t]
	\centering
	\newcommand{\tabincell}[2]{\begin{tabular}{@{}#1@{}}#2\end{tabular}}
	\renewcommand{\arraystretch}{1.5}
	\caption{Ablation studies on different pooling strategies for TALNet on MARS dataset.}
	\label{pooling}
	\begin{tabular}{|c|c|c|}
		\hline
		{\textbf{Pooling strategies}}&\textbf{Rank-1}&\textbf{mAP}\\
		\cline{2-3}
		\hline
		{Random sampling}&{77.4}&{65.6}\\
		\hline
		{Max}&{85.5}&{74.8}\\
		\hline
		{Mean}&{89.1}&{82.3}\\
		\hline
	\end{tabular}
\end{table}


The results in Table \ref{AppNet} show the influence of different appearance features on person re-identification. All experimental features are extracted from the base network followed by the appearance branch network. AppNet\_w/o global denotes only abstracting local appearance features from body parts of pedestrian, which acquires 83.5\% Rank-1 and 73.5\% mAP. AppNet\_w/o local means only extracting the discriminative global features from the human body, which achieves 84.2\% Rank-1 and 75.0\% mAP. AppNet\_ w/o GRU refers to using mean pooling operation on the frame-level global and local features, instead of the two GRU layers, which achieves 84.5\% Rank-1 and 75.4\% mAP. AppNet denotes utilizing the two GRU layers to encode the frame-level global and local features as video-level representations, which achieves the best performance in term of Rank-1 and mAP. By comparing the results of the ablation experiment, we can find that the global feature and local features are complementary to each other in person re-identification task. The joint learning of them could effectively increase the accuracy of video-based person retrieval. In addition, when comparing the results of AppNet and AppNet\_ w/o GRU, we could conclude that the GRU layers exploit temporal contextual information among video frames and promote the learning of appearance representations.

Table \ref{AttNet} reports the accuracy for each component of the attribute branch network. These comparative experiments are performed by the base network with the attribute branch network. AttNet\_w/o ST denotes the attribute branch network without the spatial attention block and the temporal-semantic context block, which directly uses the extracted features for the attribute classification. In Table \ref{AttNet},  its Rank-1 metric reaches 54.5\% and mAP is 37.9\%. AttNet\_w/o T denotes the attribute branch network without the temporal-semantic context block, which obtains 60.2\% Rank-1 and 40.6\% mAP. AttNet\_w/o C denotes the attribute branch network without the context memory unit, which achieves 62.7\% Rank-1 and 45.8\% mAP. AttNet denotes the full attribute branch network. Compared with other models in Table \ref{AttNet}, AttNet achieves the excellent performance on Rank-1 and mAP. Its Rank-1 metric is as high as 65.5\% and mAP is up to 47.9\%. Comparing the performance of AttNet\_w/o ST and AttNet\_w/o T, meanwhile, it can be seen that the spatial attention block could construct an accurate visual attention for each attribute classification and thus boost the learning of attribute representations. From the comparison between AttNet\_w/o T and AppNet, the temporal-semantic context block is able to explore the temporal context for each attribute across video frames and the semantic context among attributes simultaneously, towards improving the accuracy of attribute recognition. Moreover, by comparing the performance of AttNet\_w/o C and AppNet, the context memory unit in the temporal-semantic context block is able to drive the network to selectively concentrate on the crucial temporal-semantic context among attributes, and generate more effective attribute representation. 
	
We also compare the performance of TALNet by employing various pooling operations (\textit{e.g.}, average pooling, max pooling, and random sampling pooling) to abstract effective video-level representations. For average or max pooling operation, every eight consecutive frames in a video sequence are formed as a clip, which is fed into TALNet to extract the pedestrian feature. All the clip-level features are aggregated into the final video representation by different pooling operations. For random sampling pooling, we randomly sample eight frames from the same video sequence and pass it into TALNet to generate the video representation. The analysis results of TALNet with three pooling operations are demonstrated in Table \ref{pooling}. Rank-1 accuracy of random sampling, max and average pooling are 77.4\%, 85.5\%, and 89.1\%, respectively, and mAP are 65.6\%, 74.8\%, and 82.3\%, respectively. Compared to the random sampling and max pooling operations, the average pooling operation achieves the outstanding performance on Rank-1 and mAP. Consequently, we adopt mean pooling operation to aggregate the features of all clips of a video sequence and generate the final video representations used for pedestrian matching.

\section{Conclusions}
In this work, we propose a novel Temporal Attribute-Appearance Learning Network (TALNet) to learn discriminative and robust pedestrian representation for video-based person re-identification. The proposed TALNet jointly learns appearance and attribute representations of pedestrians, which simultaneously exploits the complementation between the two representations, hard visual attention and the temporal-semantic context for human attributes across video frames, as well as the spatial-temporal dependencies among body parts for human appearance. The appearance features extracted from both whole body and body parts offer a comprehensive description for person appearance. The exploration of hard visual attention on attributes and temporal-semantic context among attributes across video frames, results in precise learning of attributes and robust attribute representation. We conducted extensive experiments on two challenging benchmarks, \textit{i.e.}, MARS and iLIDS-VID datasets. The experimental results have shown that the proposed TALNet outperforms a wide range of state-of-the-art methods by a large margin, which validates the effectiveness of TALNet.
	
\ifCLASSOPTIONcaptionsoff
  \newpage
\fi



\bibliographystyle{IEEEtran}
\bibliography{mybib}
%
%








\end{document}